\documentclass[journal]{IEEEtran}
\IEEEoverridecommandlockouts %
\usepackage{multirow}
\usepackage{cite}
\usepackage{hyperref}
\usepackage[normalem]{ulem}
\usepackage{adjustbox,booktabs}
\usepackage{graphicx} %
\usepackage{subcaption} %
\usepackage{amsmath} %
\usepackage{amssymb}  %
\usepackage{mathrsfs}  %

\usepackage{amsthm}  %
\usepackage{circledsteps}
\usepackage{adjustbox}
\usepackage{algorithm}
\usepackage{algpseudocode}

\newtheorem{proposition}{Proposition}

\newtheorem{remark}{Remark}
\newtheorem{corollary}{Corollary}
\newtheorem{problem}{Problem}
\newtheorem{assumption}{Assumption}

\title{\LARGE \bf
Data-driven Spatial Classification using Multi-Arm Bandits for Monitoring with Energy-Constrained Mobile Robots}

\author{
Xiaoshan Lin, Siddharth Nayak, Stefano Di Cairano, Abraham P. Vinod$^{\dagger}$
\thanks{$^*$ Corresponding author.  \newline
\indent X. Lin is with the Aerospace Engineering and Mechanics department, University of Minnesota, Minneapolis, MN 55455, USA  (email: \texttt{lin00668@umn.edu}).
S. Nayak is with the Aeronautics and Astronautics department, Massachusetts Institute of Technology, Cambridge, MA 02139, USA (email: \texttt{sidnayak@mit.edu}). 
S. Di Cairano and A. Vinod are with Mitsubishi Electric Research Laboratories (MERL), Cambridge, MA 02139, USA (email: \texttt{\{dicairano,abraham.p.vinod\}@ieee.org}). 
This work was developed during the MERL internships of Lin and Nayak.
}
}

\begin{document}
\newcommand{\bbc}{\mathbb{C}}
\newcommand{\bbe}{\mathbb{E}}
\newcommand{\bbn}{\mathbb{N}}
\newcommand{\bbp}{\mathbb{P}}
\newcommand{\bbr}{\mathbb{R}}
\newcommand{\bbs}{\mathbb{S}}
\newcommand{\bbz}{\mathbb{Z}}
\newcommand{\bfc}{\mathbf{C}}
\newcommand{\bfx}{\mathbf{X}}
\newcommand{\cala}{\mathcal{A}}
\newcommand{\calb}{\mathcal{B}}
\newcommand{\calc}{\mathcal{C}}
\newcommand{\cald}{\mathcal{D}}
\newcommand{\cale}{\mathcal{E}}
\newcommand{\calf}{\mathcal{F}}
\newcommand{\calg}{\mathcal{G}}
\newcommand{\calh}{\mathcal{H}}
\newcommand{\cali}{\mathcal{I}}
\newcommand{\calj}{\mathcal{J}}
\newcommand{\calk}{\mathcal{K}}
\newcommand{\call}{\mathcal{L}}
\newcommand{\calm}{\mathcal{M}}
\newcommand{\caln}{\mathcal{N}}
\newcommand{\calo}{\mathcal{O}}
\newcommand{\calp}{\mathcal{P}}
\newcommand{\calq}{\mathcal{Q}}
\newcommand{\calr}{\mathcal{R}}
\newcommand{\cals}{\mathcal{S}}
\newcommand{\calt}{\mathcal{T}}
\newcommand{\calu}{\mathcal{U}}
\newcommand{\calv}{\mathcal{V}}
\newcommand{\calw}{\mathcal{W}}
\newcommand{\calx}{\mathcal{X}}
\newcommand{\caly}{\mathcal{Y}}
\newcommand{\calz}{\mathcal{Z}}
\newcommand{\scra}{\mathscr{A}}
\newcommand{\scrb}{\mathscr{B}}
\newcommand{\scrc}{\mathscr{C}}
\newcommand{\scrd}{\mathscr{D}}
\newcommand{\scre}{\mathscr{E}}
\newcommand{\scrg}{\mathscr{G}}
\newcommand{\scrh}{\mathscr{H}}
\newcommand{\scrl}{\mathscr{L}}
\newcommand{\scro}{\mathscr{O}}
\newcommand{\scrp}{\mathscr{P}}
\newcommand{\scrr}{\mathscr{R}}
\newcommand{\scrs}{\mathscr{S}}
\newcommand{\scru}{\mathscr{U}}
\newcommand{\scrv}{\mathscr{V}}
\newcommand{\scrw}{\mathscr{W}}
\newcommand{\scrx}{\mathscr{X}}
\newcommand{\sn}[1]{\textcolor{red}{Sid: #1}}
\newcommand{\red}[1]{\textcolor{red}{#1}}

\newcommand{\Nint}[2]{\mathbb{N}_{[#1, #2]}}

\newcommand{\nCellsToInvestigateAtEpoch}{D}
\newcommand{\keepSetI}{\mathcal{K}}
\newcommand{\rejectionSetI}{\mathcal{R}}
\newcommand{\keepSet}[1]{\keepSetI(#1)}
\newcommand{\rejectionSet}[1]{\rejectionSetI(#1)}
\newcommand{\termTime}{p_{\text{term}}}
\newcommand{\roadSet}{\scrr}
\newcommand{\obstacleSet}{\scro}
\newcommand{\candidate}{\scrc}

\maketitle
\thispagestyle{empty}
\pagestyle{empty}

\begin{abstract}
We consider the spatial classification problem for monitoring using data collected by a coordinated team of mobile robots. Such classification problems arise in several applications including search-and-rescue and precision agriculture. Specifically, we want to classify the regions of a search environment into interesting and uninteresting as quickly as possible using a team of mobile sensors and mobile charging stations. We develop a data-driven strategy that accommodates the noise in sensed data and the limited energy capacity of the sensors, and generates collision-free motion plans for the team. We propose a bi-level approach, where a high-level planner leverages a multi-armed bandit framework to determine the potential regions of interest for the drones to visit next based on the data collected online. Then, a low-level path planner based on integer programming coordinates the paths for the team to visit the determined regions subject to the physical constraints. We characterize several theoretical properties of the proposed approach, including anytime guarantees and task completion time. We show the efficacy of our approach in simulation, and further validate these observations in physical experiments using mobile robots.
\end{abstract}


\section{Introduction}
\label{sec:intro}
Monitoring extensive areas using autonomous search teams has several applications in infrastructure maintenance, search and rescue operations, and wildlife tracking~\cite{ifac,drew2021multi,queralta2020collaborative,6161683}. 
In this paper, we study the spatial classification problem that requires rapid identification of regions in a search environment containing interesting objects or phenomena. 
We tackle the spatial classification problem using a coordinated team of heterogeneous robots comprising of mobile sensors and mobile charging stations.
The mobile sensors (e.g. drones) are used for collecting data from the environment, while the mobile stations (e.g. ground vehicles) address the energy limitations of drones.
Apart from the physical constraints on the existing mobile robotic platforms like dynamics, collision-avoidance, and battery, the deployment strategies for the team must also accommodate noisy measurements arising from low-cost, low-weight onboard sensors. 
This paper, builds on our recent work~\cite{Siddharth2024data}, to propose \emph{a data-driven spatial classification algorithm that explicitly accounts for physical constraints on the robot team and noisy data collected by mobile sensors.}

Various strategies have been proposed for multi-agent monitoring~\cite{drew2021multi,queralta2020collaborative,6161683}, including submodular maximization~\cite{krause2007near}, Voronoi-based search with function approximation~\cite{bullo2009distributed, schwager2015robust, luo2019distributed}, active sensing/perception~\cite{bajcsy2018revisiting}, graph-based search~\cite{kapoutsis2017darp, best2018online}, and statistical learning~\cite{marchant2012bayesian, pmlr-v161-ghods21a}. Despite their effectiveness, many of these approaches may require perfect sensing, may lack finite-time guarantees on search performance, may assume spatial correlation, may overlook the team heterogeneity, or may ignore/relax the physical constraints on the team.

Multi-arm bandits are a special class of reinforcement learning algorithms, designed for problems where the current actions \emph{do not} impact future reward~\cite{lattimore2020bandit}.
These algorithms exhibit non-asymptotic guarantees of performance, and have been recently  applied to monitoring tasks~\cite{rolf2018successive, Hassan2022MultiSourceDetection,ifac}. 
In~\cite{rolf2018successive,Hassan2022MultiSourceDetection}, bandits-based algorithms are used to identify the top-$k$ points of interest with a probabilistic finite-time guarantee, but required prior knowledge on the number
of interesting regions. 
Our prior work~\cite{ifac} used \emph{thresholding bandits}~\cite{locatelli2016optimal,pmlr-v51-jun16,mason2020finding} and relaxed such a requirement to identify \emph{all} interesting regions. However, these approaches consider the physical limitations on the team (dynamics and energy) only as soft constraints.

Energy-aware coordinated control of heterogeneous teams of aerial and ground vehicles has gained increasing attention~\cite{queralta2020collaborative,drew2021multi,6161683,KunduIROS2021,XiaoshanRAL2022,yu2018algorithms}. For example, approaches based on satisfiability modulo theories  in~\cite{KunduIROS2021} plan energy-efficient trajectories for mobile charging stations given pre-computed trajectories of sensors. In contrast,~\cite{XiaoshanRAL2022} generates offline trajectories for both sensors and mobile charging stations simultaneously via partitioning, with the ability to modify them during online execution to handle unknown obstacles. Additionally, approaches based on traveling salesman problems~\cite{yu2018algorithms} have also been explored. 
However, these works may be limited to static goals due to the need for offline planning, or may not adapt to dynamic assignments of sub-teams formed by a ground vehicle and its associated aerial vehicles. 

The main contribution of this work is a bi-level approach that uses a combination of multi-arm bandits and optimization to address the data-driven spatial classification problem using a heterogeneous team. We extend our recent work~\cite{Siddharth2024data} to: 1) incorporation of additional constraints into the low-level planner, and 2) online adjustment of computed paths to guarantee collision avoidance using linear assignment, and 3) an experimental validation of the proposed approach in a search-and-rescue application using drones with vision-based sensing and ground robots. We also provide theoretical guarantees for the proposed approach, including anytime guarantees (the algorithm provides a useful result, even when terminated prematurely), and finite upper bounds on task completion time.

\emph{Notation:} $|\cals|$ is the cardinality of a set $\cals$ and $\Nint{a}{b}$ is the set of natural numbers between (and including) $a,b\in \bbn$, $a \leq b$.

\section{Problem statement}

\emph{Search environment:} We define the search environment as a set $\calg$ of grid cells. Within $\calg$, let $\obstacleSet\subset\calg$ be the known set of cells that are occupied by obstacles or no-fly zones, and $\cali\subseteq\calg\setminus\obstacleSet$ be the \emph{a priori} unknown set of \emph{interesting cells} that are occupied by objects/phenomena of interest. The objective of the spatial classification problem is to identify $\cali$ as soon as possible using data collected online by a team.

\emph{Search team:} We define the search team as comprising of  $N_d\in\mathbb{N}$ mobile sensors (e.g., drones) and $N_c\in\mathbb{N}$ charging stations (e.g., autonomous trucks). For simplicity, we will refer to the mobile sensors as \emph{sensors} and the charging stations as \emph{stations} throughout the rest of the paper. 
\begin{itemize}
    \item \emph{Sensors.} The sensors are responsible for monitoring the search environment and collecting data. They are energy-limited and must recharge by rendezvous with a station every $T_d\in\mathbb{N}$ moves. 
    At each move, the sensors can move to their neighbouring cells in all cardinal directions and  all diagonal directions  (like a king piece in chess).
    For any cell $l\in\calg\setminus\obstacleSet$, $\caln(l)$ defines the set of neighboring cells that a sensor can move from $l$, and $\caln(l)\subseteq\calg\setminus\obstacleSet$.
    \item \emph{Stations.} Upon rendezvous with a sensor, a station recharges its  battery.  
    We assume that the stations have sufficient energy for the mission and do not require recharging themselves, similarly to~\cite{XiaoshanRAL2022,yu2018algorithms,Siddharth2024data}. 
    The stations move as the sensors, but their workspace may be more restricted (e.g. 
    stations can only move on the roads in a search-and-rescue scenario). 
    We define \emph{station-admissible set} $\roadSet \subseteq\calg\setminus\obstacleSet$ as the set of grid cells that can be visited by a station. 
    For any cell $l\in\roadSet$,  the set of neighboring cells that a station can move is $\caln_c(l)\triangleq\roadSet\cap\caln(l)$.
\end{itemize}
We model the difference in agility between sensors and stations by their speeds $v_d$ and $v_c$ respectively, where we assume that the ratio of their speeds $\gamma_{dc}=\frac{v_d}{v_c} \in \bbn$ is a positive integer. 
Consequently, the stations move $T_c$ cells by the time the sensors move $T_d = T_c \gamma_{dc}\geq T_c$ cells.

\emph{Sensing:} 
Let $\candidate=\calg\setminus\obstacleSet$ be  the set of candidate cells that may potentially be ``interesting'', i.e., it contains the search objective with $\cali\subseteq\candidate$. 
When a sensor visits a cell in $\candidate$, it receives noisy, binary measurements (data) of whether that cell is interesting or not. 
Formally, every cell $l\in\candidate$ has a corresponding Bernoulli random variable $\nu_l$ with \emph{a priori} unknown mean $\mu_l$, and each cell visit generates $B\in\bbn$ realizations of $\nu_l$, where $B$ is the sample batch size.
The mean $\mu_l$ 
may be influenced by the underlying spatial distribution of the interesting cells as well as the imperfections of the noisy sensors and the perception algorithms used by the team.
We ignore the possible variation in measurement noise characteristics among the sensors in the team, and assume that $\nu_{l_1}$ and $\nu_{l_2}$ are independent for any $l_1,l_2\in\candidate$. 

\emph{Labeling error criterion}: For a user-defined threshold $\theta\in (0,1)$, we formally define $\cali$ as
\begin{align}
  \cali =\cals_{\theta} \triangleq \{l\in\candidate: \mu_l\geq\theta\}\label{eq:cals_theta}.
\end{align}
From the definition of $\mu_l$, $\cals_{\theta}$ are the cells with a likelihood of at least $\theta$ of being sensed as interesting.
Thus, the spatial classification problem of interest is to identify $\cals_\theta$ using noisy measurements collected by the team.

We assume tolerance to labeling errors for approximating $\cals_\theta$ using noisy data in finite time, similarly to~\cite{ifac,Siddharth2024data}.
\begin{assumption}
We ignore the labeling error for all cells $l\in\calg$ with $\mu_l \in (\theta-\epsilon, \theta+\epsilon)$ for some small tolerance $\epsilon > 0$. \label{assum:labeling_error}
\end{assumption}
Under Assumption~\ref{assum:labeling_error}, $\cals_\theta$ may be approximated by a (keep) set $\keepSetI\subseteq\calg$ such that $\cals_{\theta+\epsilon}\subseteq \keepSetI\subseteq\cals_{\theta-\epsilon}$. 
For a given \emph{labeling error probability} $\delta\in(0,1)$, the spatial classification problem may be (approximately) solved by identifying $\keepSetI$ that satisfies
\begin{align}
\mathbb{P}\left[{\left({\cals_{\theta+\epsilon}\setminus\keepSetI}\right)\cup\left({\keepSetI\setminus\cals_{\theta-\epsilon}}\right)
= \emptyset}\right] \geq 1-\delta.\label{eq:labeling_error}
\end{align}%
We refer to \eqref{eq:labeling_error} as the \emph{labeling error} criterion, which ensures $\cals_{\theta+\epsilon}\subseteq \keepSetI\subseteq\cals_{\theta-\epsilon}$ by requiring that $\keepSetI$ includes almost all interesting cells ($\mu_l\geq\theta+\epsilon$) and excludes almost all uninteresting cells ($\mu_l\leq\theta-\epsilon$) with probability $1-\delta$.
 
We now state the two problems tackled by this paper:
\begin{problem}
    Design a data-driven algorithm for the search team that terminates in finite time, upon termination meets the labeling error criterion  \eqref{eq:labeling_error}, and ensures that the motion and energy constraints on the team are satisfied. \label{prob_st:main}
\end{problem}

\begin{problem}
    For the algorithm solving  Problem~\ref{prob_st:main},  
    determine upper bounds on the time to terminate the search.
    \label{prob_st:main_bounds}
\end{problem}

\section{Methodology}
\begin{figure}[t]
        \centering
        \includegraphics[height=0.7\linewidth]{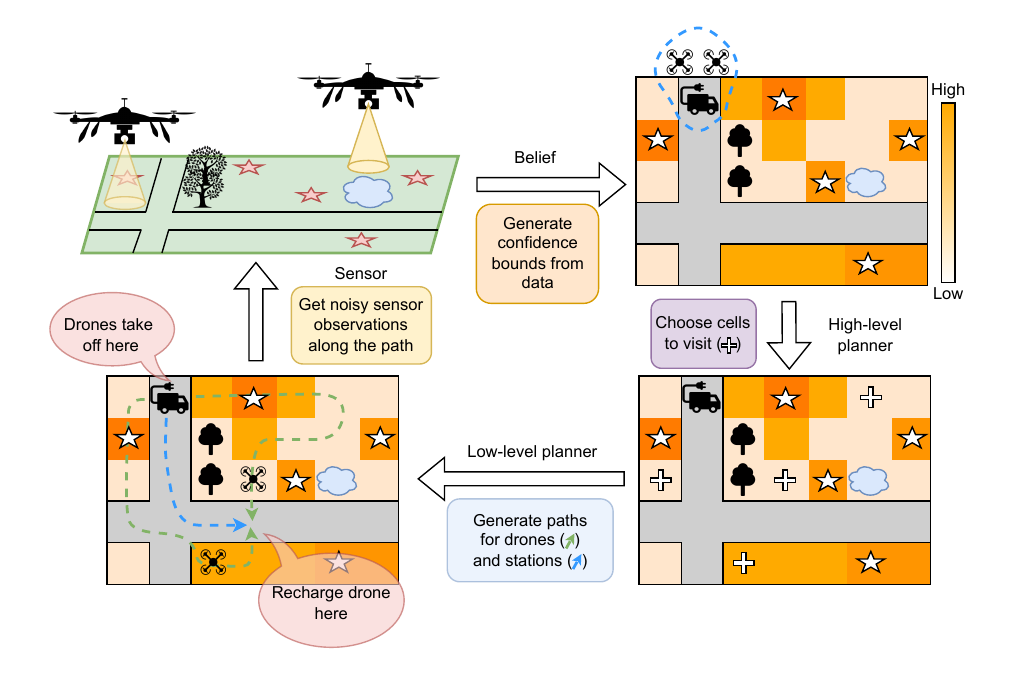}
        \caption{Data-driven multi-agent search under noisy observations. 
        A high-level planner uses the available measurements of cells to determine the confidences of an unclassified cell to be interesting, and determines the cells to visit next. 
        A low-level planner computes trajectories for the agents subject to various motion constraints arising from dynamics as well as energy limitations.
        Altogether, the proposed approach guarantees safe, data-driven sensor deployment to address Problem~\ref{prob_st:main}.}
        \label{fig:illustration_method}
    \end{figure}

    We address Problem~\ref{prob_st:main} by an iterative algorithm that uses a bi-level approach, as illustrated in Figure~\ref{fig:illustration_method}. 
    We use \emph{epoch} $p\in \bbn$ to denote time in a slower time scale used by a high-level planner, as compared to \emph{time step} $t\in \bbn$, which denotes time in a faster time scale used by a low-level planner. 
    The high-level planner identifies a set of \emph{epoch goals} based on available (noisy) data collected online. 
    Epoch goals are the cells selected for the team to visit during the current decision epoch to gather additional measurements.
    We use the multi-arm bandits framework to design the high-level planner. Given a set of epoch goals, the low-level planner generates motion plans for the team to visit the epoch goals and collect the new measurements. 
    The two-staged algorithm continues until a termination criterion is met. 
    In this section, we detail the planners, and describe the termination criterion.

\subsection{High-level Planner: Bandit-based decision making}
\label{sub:high_level}
We propose a high-level planner based on the bandit monitoring algorithm in our prior work~\cite{ifac}. Specifically, the spatial classification problem is cast as a $|\candidate|$-armed bandit problem, where the bandit arms are the cells.
We design the high-level planner as a sequential decision maker that processes the collected information to decide which cells must be visited at each epoch.
We use \emph{upper confidence bounds} typical of bandit-based algorithms~\cite{locatelli2016optimal,mason2020finding,pmlr-v51-jun16,lattimore2020bandit} to sample the unclassified cells that are ``most likely" to be interesting --- the epoch goals.
Additionally, the high-level planner maintains the \emph{keep set} $\keepSetI$ and the \emph{reject  set} $\rejectionSetI$, the set of cells classified as interesting and uninteresting, respectively.

Let $\calh_l(p)$ be the history of measurements taken at cell $l\in\candidate$ collected by the search team until epoch $p$, and define $\calh(p)={\{\calh_l(p)\}}_{l\in\candidate}$. Then, at epoch $p$, we choose $\nCellsToInvestigateAtEpoch$ (typically, $\nCellsToInvestigateAtEpoch\geq N_d$)
distinct cells that achieve the top-$D$ values of a 
function $J:\candidate\times\mathbb{N}\times(0,1)\to\mathbb{R}$,
\begingroup\makeatletter\def\f@size{9.5}\check@mathfonts
\begin{subequations}
\begin{align}
    J(l,p,\delta)&=\hat{\mu}_l(p) + U_l(p, \delta),\qquad
    \hat{\mu}_l(p) = \frac{\sum_{h\in \calh_l(p)} h}{|\calh_l(p)|},\\
    U_l(p,\delta) &= 2\sqrt{\frac{2\log(\log_2(2|\calh_l(p)|))+\log\left({12|\candidate|/\delta}\right)}{2|\calh_l(p)|}},\label{eq:acq_fun_U}
\end{align}\label{eq:acq_fun}%
\end{subequations}%
\endgroup
with $\hat{\mu}_l(p)=U_l(p,\delta)=\infty$, whenever $\calh_l(p)=\emptyset$. Here, $\delta\in(0,1)$ is a given (small) labeling error probability. After getting the data for the current epoch, we update the sets
$\keepSet{p+1}$ and $\rejectionSet{p+1}$ as follows,
\begingroup\makeatletter\def\f@size{9.5}\check@mathfonts
\begin{subequations}
\begin{align}
    \keepSet{p+1} &= \{l\in \candidate : \hat{\mu}_l(p+1) - U_l(p+1,
\delta) \geq \theta -\epsilon \},\\
    \rejectionSet{p+1} &=\{l\in \candidate :  \hat{\mu}_l(p+1) + U_l(p+1,
\delta) \leq \theta + \epsilon \}.
\end{align}\label{eq:update_sets}%
\end{subequations}%
\endgroup
Since
$U_l(p,\delta)$ is a non-increasing function
of $|\calh_l|$ and $|\calh_l|$ is a non-decreasing function in $p$, the sets $\keepSetI$ and $\rejectionSetI$ are
monotonic in $p$ with $|\keepSetI|$ and $|\rejectionSetI|$ non-decreasing in $p$.
Here, \eqref{eq:acq_fun} and
\eqref{eq:update_sets} are motivated by the desire to obtain \emph{anytime guarantees} (see Section \ref{sec:theoretical_guarantees}) along with ensuring a low classification time. 

\begin{figure}
    \centering
    \includegraphics[width=1\linewidth]{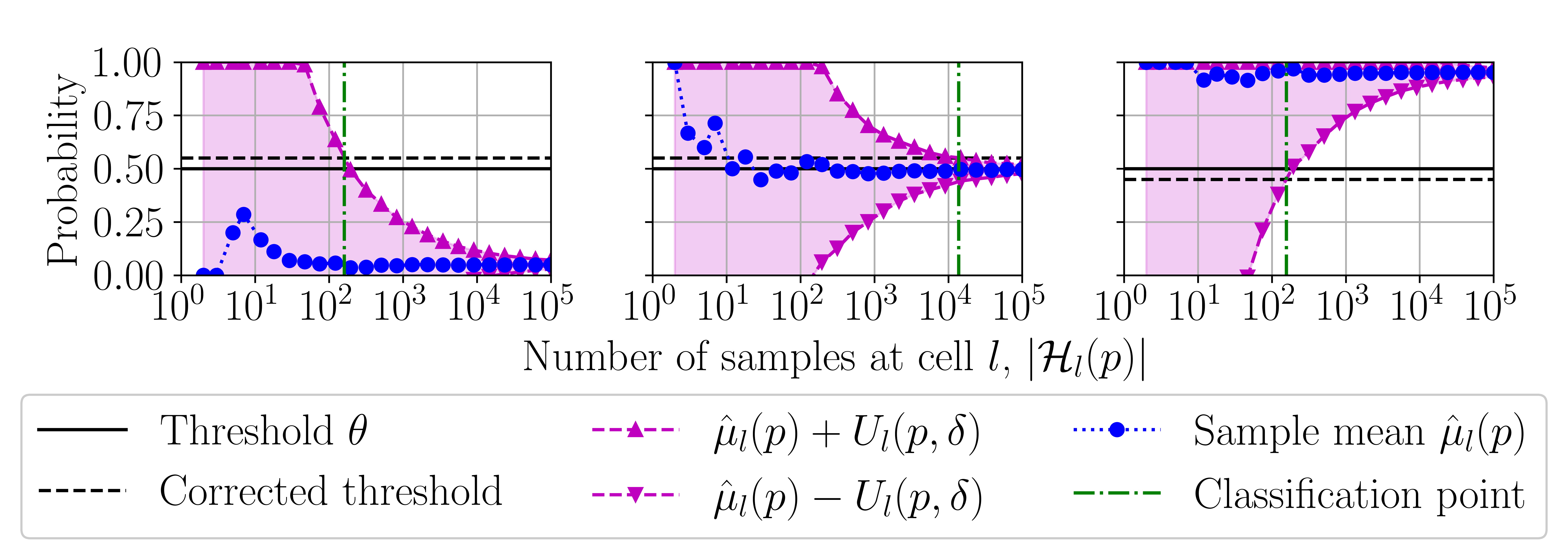}
    \caption{Sample mean and bounds \eqref{eq:update_sets} for $\mu_l\in\{0.05, 0.5, 0.95\}$ with $\theta=0.5$, $\epsilon=0.05$, $|\calg|=100$, and $\delta=0.0001$. Here, \eqref{eq:update_sets} classifies $\mu_l\in\{0.05,0.5\}$ (left, middle) as uninteresting $(l\in\calr)$, and $\mu_l=0.95$ (right) as interesting $(l\in\calk)$.}
    \label{fig:bounds}
\end{figure}

Figure~\ref{fig:bounds} illustrates how \eqref{eq:update_sets} classify cells for three Bernoulli variables $\nu_l$ with mean $\mu_l\in\{0.05, 0.5, 0.95\}$ respectively. Cells with large $|\mu_l-\theta|$ need fewer samples for classification.

Algorithm~\ref{algo:high_level} summarizes the proposed approach to address Problem~\ref{prob_st:main}, also shown in  Figure~\ref{fig:illustration_method}. 
Step~\ref{step:high_level_epoch_goal} of Algorithm~\ref{algo:high_level} identifies the epoch goals for the search team using \eqref{eq:acq_fun}.
Step~\ref{step:high_level_low_level} uses a low-level planner to design a deployment for the search team to visit at all epoch goals, which we describe next.

    \begin{algorithm}[t]
        \caption{Bandits-based, autonomous spatial classification}\label{algo:high_level}
        \begin{algorithmic}[1]
        \Require Set of grid cells $\calg$, threshold $\theta\in(0,1)$, labeling error tolerance $\epsilon >0$, labelling error probability $\delta \in (0,1)$, number of epoch goals $\nCellsToInvestigateAtEpoch \in \mathbb{N}$, sample batch size $B\in\bbn$, obstacle set $\obstacleSet$, station admissible set $\roadSet$
        \Ensure $\{\keepSet{p}\}_{p\in\bbn}$, sequence of (keep) sets
        \State Initialize $p\gets0$,  $\keepSet{p}\gets\emptyset$, and $\rejectionSet{p}\gets\emptyset$
        
        \While{$\rejectionSet{p} \cup \keepSet{p} \neq \candidate $}
        
        \State Define $\cale_p$ as the top $\nCellsToInvestigateAtEpoch$ elements in the list of unclassified cells $l\in\candidate\setminus(\keepSet{p}\cup\rejectionSet{p})$, sorted in descending order based on $J(l,p,\delta)$ \eqref{eq:acq_fun}\label{step:high_level_epoch_goal}

        \State Deploy the search team to visit cells in $\cale_p$ while avoiding $\obstacleSet$ by using paths generated by the low-level planner (Algorithm \ref{algo:low_level_ip}), take measurements along the way, and update history $\calh(p+1)$\label{step:high_level_low_level}

        \State Update sets of classified cells $\keepSet{p+1}$ and $\rejectionSet{p+1}$ based on \eqref{eq:update_sets} using $\calh(p+1)$, $\candidate$, $\theta$, $\delta$, and $\epsilon$ 

        \State Increment epoch $p\gets p+1$
        \EndWhile
        \end{algorithmic}
    \end{algorithm}
    
    \subsection{Low-level coordination: Integer Program-based solution} 

    The optimization-based planner determines the deployment plan for the team based on the epoch goals identified by the high-level planner.
    Since the team may not be able to visit all cells in $\cale_p$ using a single charge of their batteries, we divide the deployment into \emph{sensing cycles}.
    Each sensing cycle is such that the sensors always start and end on a station, the sensing cycles are of length  $T_d$, and the paths of the sensors and the stations in sensing cycles satisfy their respective motion constraints. 
    Subsequently, the optimization-based low-level planner solves the following problem at each epoch $p$,
\begingroup\makeatletter\def\f@size{9.5}\check@mathfonts
\begin{subequations}\label{prob:informal_ip}
\begin{align}
        \text{min.} 
        &\quad \text{Number of sensing cycles}, \\
         \text{s. t.}
         &\quad \text{Search team respects motion constraints}\label{eq:informal_ip_motion},\\
         &\quad \text{Search team visits all cells in $\cale_p$},\label{eq:informal_ip_visit_e}\\
         &\quad \text{Search team never visits $\obstacleSet$, and stations stay in $\roadSet$},\label{eq:informal_ip_visit_o}\\
         &\quad\text{Sensors never run out of battery}.\label{eq:informal_ip_no_battery_run_out}
\end{align}%
\end{subequations}%
\endgroup

We formulate the optimization problem \eqref{prob:informal_ip} as an integer program (IP), and solve it using off-the-shelf solvers~\cite{GUROBI}.
    We use the following set of binary decision variables in the IP formulation: 
    $x_{ijkl}, y_{abkl}, \lambda_{k} \in \{0,1\}$ for every
    \begingroup\makeatletter\def\f@size{9}\check@mathfonts
\begin{align}
    \begin{array}{lll}
        \text{(Sensor/station index):} & i\in\cald\triangleq\Nint{1}{N_d},& a\in \calc\triangleq\Nint{1}{N_c},\\
        \text{(Sensing cycle time):} &j\in \calt_d\triangleq\Nint{0}{T_d-1},& b\in \calt_c\triangleq\Nint{0}{T_c-1},\\ 
        \text{(Sensing cycle count):} & k\in \calt_s\triangleq\Nint{0}{K-1},\\%& c\in \calk,\\
        \text{(Grid cell location):} & l\in \calg\triangleq\Nint{1}{|\calg|},%
    \end{array} \label{eq:bin_var_indices_dfn}
    \raisetag{14pt}
    \end{align}
    \endgroup
    where 
    $x_{ijkl}=1$ if sensor $i$ is at cell $l$ at time step $j$ in the sensing cycle $k$, $x_{ijkl}=0$ otherwise; $y_{abkl}=1$ if station $a$ is at cell $l$ at time step $b$ in the sensing cycle $k$, $y_{abkl}=0$ otherwise; and, $\lambda_k=1$ if sensing cycle $k$ is necessary for the search team to solve \eqref{prob:informal_ip}, $\lambda_k=0$ otherwise. Here, $K$ is a user-specified upper bound on the number of sensing cycles. 
    In what follows, the constraints are enforced for all indices $i,j,k,l,a$, and  $b$ as described in \eqref{eq:bin_var_indices_dfn}, unless stated otherwise.

    To simplify notation and discussion, \eqref{eq:bin_var_indices_dfn} associates a binary variable for every $l\in\calg$ instead of associating only binary variables within $\candidate$ for the sensors and within $\roadSet$ for the stations.
    The number of variables in the resulting IP for each epoch $p$ is $K(1 + |\calg|(N_dT_d + N_cT_c))$. 
    
    While an integer program formulation of \eqref{prob:informal_ip} may require non-trivial computational effort (especially for large teams and large search environments), we remind the reader that our setup has relatively permissible bounds on solve time for \eqref{prob:informal_ip}, compared to traditional motion planning problems. Specifically, at each epoch, all sensors will be charging their batteries after their rendezvous with a station. 
    Thus, it suffices to solve \eqref{prob:informal_ip} within the time to recharge the batteries of sensors, which may be in the order of tens of minutes.
    
    \emph{Motion constraints:} 
    The following constraints enforce \eqref{eq:informal_ip_motion},\vspace*{-1.5em}
    \begingroup\makeatletter\def\f@size{9}\check@mathfonts\begin{subequations}
    \begin{align}
        x_{i00l}=1\text{ and } y_{a00d}&=1, &&\forall l\in \calx_0, d\in \caly_0,\label{eq:motion_constraint_initial_constraint}\\
        \sum\nolimits_{m\in\caln(l)}x_{i(j-1)km}&\geq x_{ijkl},&&\forall j\in\Nint{1}{T_d-1}\label{eq:motion_constraint_x}\\
        \sum\nolimits_{m\in{\caln_c(l)}}y_{a(b-1)km}&\geq y_{abkl},&&\forall b\in\Nint{1}{T_c-1}\label{eq:motion_constraint_y}\\
        \sum\nolimits_{a\in\calc} y_{abkl} &\leq 1\label{eq:motion_constraint_no_visit_no_collision_y}\\
        \sum\nolimits_{i\in\cald} x_{ijkl} &\leq 1\label{eq:motion_constraint_no_visit_no_collision_x}, && \forall j\in\Nint{1}{T_d - 2}\\
         \sum\nolimits_{l\in\calg} x_{ijkl} =\sum\nolimits_{l\in\calg} y_{abkl} &=\lambda_k\label{eq:motion_constraint_no_visit_more_cells},\\
        \lambda_{k} &\geq \lambda_{k+1},&&\forall k\in\calt_s. \label{eq:motion_constraint_lambda}
    \end{align}\label{eq:motion_constraint}%
    \end{subequations}%
    \endgroup
    Constraint \eqref{eq:motion_constraint_initial_constraint} sets the sensor and the station locations to the team configuration $(\calx_0,\caly_0)$ at epoch $p$, where $\calx_0, \caly_0\subset\calg$ is the set of cells occupied by the sensors and stations respectively at epoch $p$. 
    Constraints  \eqref{eq:motion_constraint_x} and \eqref{eq:motion_constraint_y} enforce the transitions for sensors and stations from their neighbors to their current cells respectively.
    Constraint  \eqref{eq:motion_constraint_no_visit_no_collision_y} requires no two stations to occupy the same cell at any time to avoid collisions. Constraint  \eqref{eq:motion_constraint_no_visit_no_collision_x} enforces a similar collision avoidance among sensors, but is relaxed at the beginning and the end of epochs to facilitate multiple sensors to rendezvous with the same station.
    Alternatively, we may guarantee collision avoidance among sensors by requiring them to fly at different altitudes~\cite{Siddharth2024data}.
    Constraint  \eqref{eq:motion_constraint_no_visit_more_cells} link the binary variables corresponding to sensing cycles $\lambda_k$ to the team paths. 
    Specifically, $\lambda_k=0$ requires the team to visit no cells in $\calg$ during the sensing cycle $k$, i.e., the path of the team terminates prior to sensing cycle $k$. 
    On the other hand, when $\lambda_k=1$, each sensor and station visit exactly one cell in $\calg$ at each time step in the sensing cycle $k$. 
    Constraint \eqref{eq:motion_constraint_lambda} encodes the temporal constraint among sensing cycles, i.e.,  $\lambda_{m+1}=1$ for any $m\in\calt_s$ implies that $\lambda_k=1,\forall k\in\Nint{0}{m}$.
   
    \emph{Visit constraints}: 
    The following constraints enforce \eqref{eq:informal_ip_visit_e} and \eqref{eq:informal_ip_visit_o} by requiring that every cell $m\in\cale_p$ is visited by some sensors at some time step in some sensing cycle, the team (both sensors and stations) never visits any cell in $\obstacleSet$, and the stations only visit cells in $\roadSet$, \begingroup\makeatletter\def\f@size{9}\check@mathfonts
   \begin{subequations}
        \begin{align}
        \sum\nolimits_{(i,j,k)\in\cald\times\calt_d\times\calt_s}x_{ijkl} &\geq1,&& \forall l\in \cale_p,\label{eq:visitation_visit_cale_p} \\
        x_{ijkl}=y_{abkl}&=0,
        &&\forall l \in \obstacleSet,\label{eq:visitation_no_visit_o}\\
        y_{abkl}&=0.
        &&\forall l \in \calg\setminus\roadSet.\label{eq:visitation_no_visit_h}
        \end{align}\label{eq:visitation}%
    \end{subequations}%
    \endgroup
   \emph{Battery constraints:} The following constraints enforce \eqref{eq:informal_ip_no_battery_run_out},%
    \begingroup\makeatletter\def\f@size{9}\check@mathfonts
    \vspace{-1.5mm}
    \begin{subequations}
    \begin{align}
        \sum\nolimits_{(j,l)\in \calt_d\times\calg}x_{ijkl} &\leq T_d, \quad \sum\nolimits_{(b,l)\in \calt_c\times\calg}y_{abkl} \leq T_c, \label{eq:battery_const_length_x} \\
        |x_{i0(k+1)l} - x_{i(T_d-1)kl}| &\leq 2(1 - \lambda_{k+1}),\quad \forall k\in\Nint{1}{K-1}, \label{eq:battery_const_same_station_x} \\
        |y_{a0(k+1)l} - y_{a(T_c-1)kl}| &\leq 2(1 - \lambda_{k+1}),\quad \forall k\in\Nint{1}{K-1}, \label{eq:battery_const_same_station_y} \\
        x_{i0kl} \leq \sum\nolimits_{a\in\calc} y_{a0kl}, & \quad x_{i(T_d-1)kl} \leq \sum\nolimits_{a\in\calc} y_{a(T_c-1)kl}\label{eq:battery_const_end_on_station}
    \end{align}\label{eq:battery_const}%
\end{subequations}%
\endgroup 
Constraint \eqref{eq:battery_const_length_x} requires the path lengths of sensors and stations to satisfy the energy and velocity constraints. 
Constraints \eqref{eq:battery_const_same_station_x} and \eqref{eq:battery_const_same_station_y} require the positions of sensors and stations to remain unchanged between sensing cycles, and are trivially satisfied when $\lambda_{k+1}=0$. 
Constraints \eqref{eq:battery_const_end_on_station} require every sensor to start and end its path within a sensing cycle on a cell occupied by one of the stations. 

Algorithm~\ref{algo:low_level_ip} summarizes the IP-based low-level planner. 
We can also extend the integer program \eqref{prob:ip} to impose additional constraints on the team. For example, 
 \begin{align}
\sum\nolimits_{i\in\cald} x_{i0kl} \leq  C,\quad \sum\nolimits_{i\in\cald} x_{iT_dkl} \leq C \label{eq:no_more_than_c_drones_on_station}
\end{align}
requires that the number of sensors rendezvous on each station does not exceed a pre-specified limit $C\in\bbn$. Such constraints may be motivated by resource constraints, such as the number of chargers on each station being limited to $C$.

\subsection{Collision avoidance for the sensors}

While \eqref{eq:motion_constraint_no_visit_no_collision_y} and \eqref{eq:motion_constraint_no_visit_no_collision_x} ensure that no sensors and stations occupy the same cell for collision avoidance purposes, we may still have collisions during transitions. 
We avoid such collisions via a linear assignment-based reassignment of the waypoints to the robots, similarly to~\cite{michael2008distributed}.
Given the paths of the $N_d$ sensors of any sensing cycle $k$, let $\calt_{jk}$ and $\calt_{(j+1)k}$ be the set of cell locations the sensors will visit at time $j$ and $j+1$ respectively. The linear assignment problem amounts to finding a bijection $f: \calt_{jk} \rightarrow \calt_{(j+1)k}$ that minimizes the cost function $\sum_{l \in \calt_{jk}}{C(l, f(l))}$, where $C(l, f(l))$ denotes the Euclidean distance between cell location $l$ and $f(l)$. 
Recall that linear assignment problems are a special class of integer linear program that admit polynomial time solutions~\cite{BurkardAssignment}.
The proposed reassignment ensures that the new paths for $N_d$ sensors are corrected simultaneously, the resulting paths have minimal interaction in the continuous space, and that the sensors collectively visits the same set of waypoints as if executing the original paths.
Similar reassignment strategy may be used to ensure collision avoidance among stations too.

    \begin{algorithm}[t]
        \caption{Low-level planner: Integer program}\label{algo:low_level_ip}
        \begin{algorithmic}[1]
        \Require Epoch goals  $\cale_p$ and search team configuration $(\calx_0,\caly_0)$ at epoch $p$, obstacle set $\obstacleSet$, station admissible set $\roadSet$, search team parameters $(T_d, T_c,N_d,N_c,K,\calg)$
        \Ensure Paths for the search team
        \State Compute $x_{ijkl}^\ast, y_{abkl}^\ast, \lambda_k^\ast$ by solving the integer program,
        \begin{align}
            \begin{array}{rl}
                \text{minimize} 
                &\quad \sum\nolimits_{k\in\Nint{1}{K}} \lambda_k, \\
                 \text{subject to}
                 &\quad \eqref{eq:motion_constraint}, \eqref{eq:visitation}\text{, and }\eqref{eq:battery_const}.
            \end{array}\label{prob:ip}
        \end{align}
        
        \State Compute paths for the sensors and the stations by extracting all indices $(i,j,k,l)$ and $(a,b,c,d)$ such that $x_{ijkl}^\ast=1$ and $y_{abcd}^\ast=1$, and creating a sequence of cells to traverse through indices  $i,j,k,l,a,b,c,d$
        \end{algorithmic}
    \end{algorithm}

\section{Theoretical guarantees}
\label{sec:theoretical_guarantees}

We now address Problem~\ref{prob_st:main_bounds}, and provide correctness and completion time guarantees for Algorithm~\ref{algo:high_level}, similarly to existing pure exploration literature~\cite{locatelli2016optimal,pmlr-v51-jun16,lattimore2020bandit}.
\begin{proposition}[\textsc{Anytime algorithm}]\label{prop:anytime}
At any epoch $p\in\bbn$, the sets $\keepSet{p}$ and $\rejectionSet{p}$ in 
Algorithm~\ref{algo:high_level} satisfy $\keepSet{p}\subseteq\cals_{\theta-\epsilon}$ and $\rejectionSet{p}\subseteq\candidate\setminus\cals_{\theta+\epsilon}$, with probability of at least $1-\delta$. 
\end{proposition}
See Appendix~\ref{app:anytime} for the proof.
The key takeaway from Proposition~\ref{prop:anytime} is that the keep set $\keepSet{p}$ computed by  Algorithm~\ref{algo:high_level} always inner-approximates $\cals_{\theta-\epsilon}$. Consequently, Algorithm~\ref{algo:high_level} may be prematurely terminated if required, and the intermediate solution $\keepSet{p}$ and $\rejectionSet{p}$ will contain only interesting and uninteresting cells respectively (up to the  tolerance $\epsilon$), with high probability. 

\begin{proposition}[\textsc{Finite time guarantees}]\label{prop:finite_time}
    Algorithm~\ref{algo:high_level}
        terminates within $P^{\max}\in\bbn$ epochs with probability of at least $1-\delta$, and satisfies the labeling error
criterion \eqref{eq:labeling_error} where 
    \begin{align}
        P^\text{max}&=\frac{1}{\nCellsToInvestigateAtEpoch}\sum\nolimits_{l\in\candidate\setminus
\scrd_\Delta} P_l + \max\nolimits_{l\in\scrd_\Delta} P_l,\label{eq:T_upper_bound}\\
        P_l &=\frac{16}{B\Delta_l^2}
            \log\left({
                4\sqrt{\frac{3|\candidate|}{\delta}}
                \log\left({
                    \frac{192}{\Delta_l^2}\sqrt{\frac{3|\candidate|}{\delta}}
                }\right)
            }\right),\label{eq:P_l_defn}\\
        \Delta_l &= |\mu_l-\theta| + \epsilon,\label{eq:delta_defn}
    \end{align}
where $P_l,\Delta_l$ is defined for every $l\in\candidate$, and $\scrd_\Delta$ is the union of the cell with the smallest $\Delta_l$ among all cells $l\in
\candidate$ and a set of $D-1$ cells with the
largest $\Delta_l$ among all cells $l\in
\candidate$.
\end{proposition}
See Appendix~\ref{app:finite_time} for the proof. 
By Proposition~\ref{prop:finite_time}, Algorithm~\ref{algo:high_level} terminates at some $p_{\text{term}}\leq P^{\max}$ epochs, and returns $\keepSet{p_{\text{term}}}$ that satisfies the labeling error criterion \eqref{eq:labeling_error}. Specifically,  $\keepSet{p_{\text{term}}}$ outer-approximates $\cals_{\theta+\epsilon}$ and inner-approximates $\cals_{\theta-\epsilon}$ with high probability.
Also, Algorithm~\ref{algo:high_level} terminates faster when $\Delta_l$ increases ($\mu_l$ is far away from $\theta$), and the dependence of $|\candidate|$ and $\delta$ on $P^\text{max}$ is sub-linear.
\begin{corollary}\label{corr:low_level_bound}
    Given an environment and a team, let $K^{\max}$ be the maximum number of sensing cycles needed by Algorithm~\ref{algo:low_level_ip} to cover any set of epoch goals starting from any configuration of the team. Then, Algorithm~\ref{algo:high_level} with Algorithm~\ref{algo:low_level_ip} as the low-level planner terminates while satisfying \eqref{eq:labeling_error} in not more than $P^{\max}K^{\max}T_d$ time steps with probability of at least $1-\delta$.
\end{corollary}
Corollary~\ref{corr:low_level_bound} bounds the time steps $t$ required by a team that are deployed using Algorithms~\ref{algo:high_level} and~\ref{algo:low_level_ip} to complete the search, and satisfy \eqref{eq:labeling_error}. 
Determining $K^{\max}$ involves a min-max computation, and while its solution always exists and is finite, it may be hard to find. However, one can obtain reasonable estimates of $K^{\max}$ via Monte-Carlo simulations.

\section{Hardware-based validation}

\begin{figure}[t]
    \centering
    \includegraphics[width=1\linewidth]{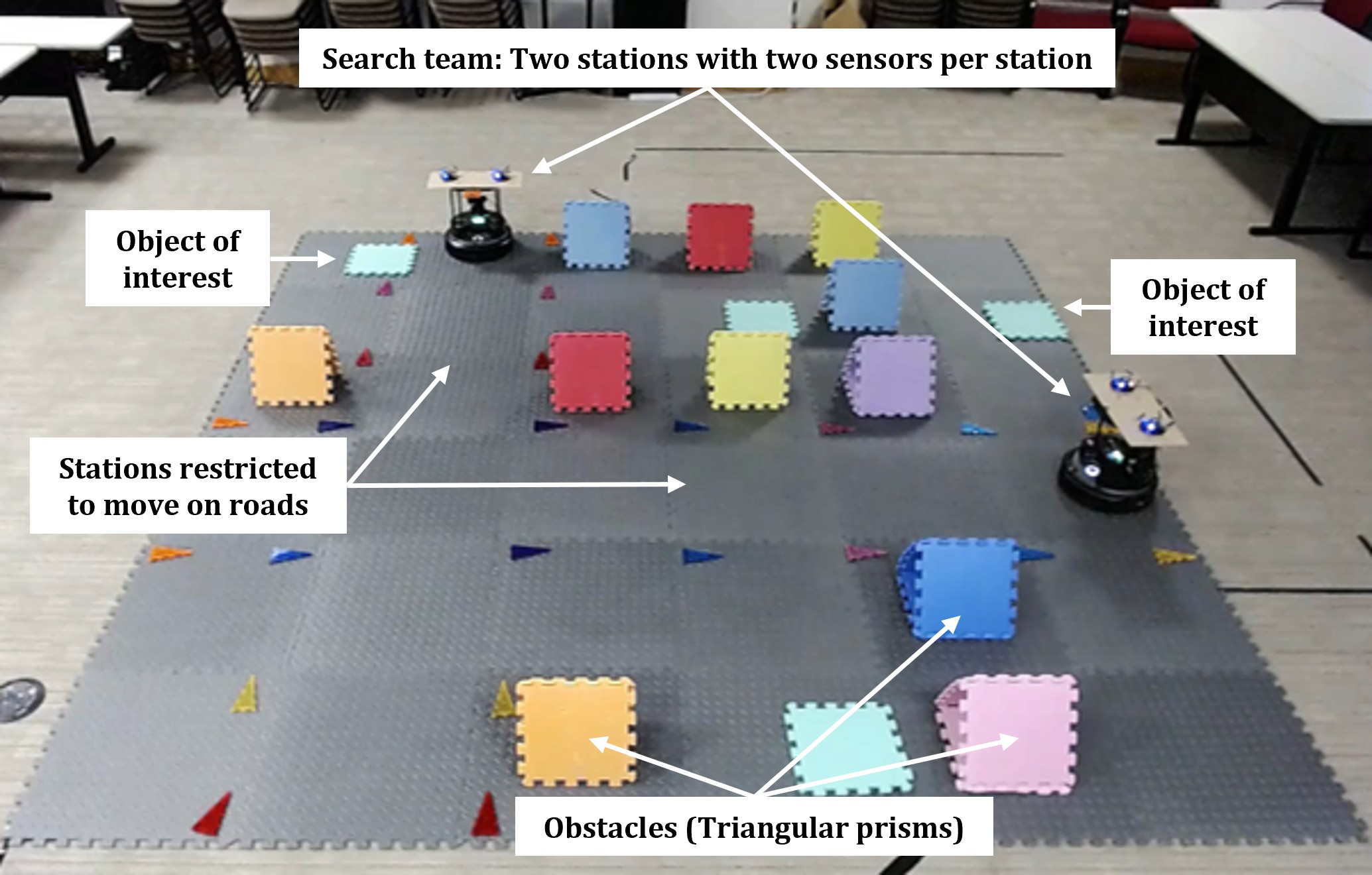}
    \caption{Snapshot of the hardware experiment with two Turtlebot4 robots as stations and four Crazyflie drones as sensors. The colored triangular prisms represent the obstacles or no-fly zones for the sensors. The second column and the third row (delimited by small triangular tiles) represent the roads where the stations must stay. The four green tiles represent the interesting cells, i.e., they contain the search objective. All other cells are considered uninteresting.
    See video of the physical experiment at \href{https://youtu.be/gzulpOcVYzg}{https://youtu.be/gzulpOcVYzg}.}
\label{fig:snapshot}
\end{figure}

We validate the proposed approach using a team comprised of four drones and two ground robots in a $3.6\times 3.6$ meter workspace partitioned into a $6\times 6$ grid environment representing a search-and-rescue application. Figure~\ref{fig:snapshot} shows the experiment setup, which includes $11$ obstacles or no-fly zones (colored triangular prisms), $4$ interesting regions (green tiles), $2$ roads (delimited by triangular tiles), and $10$ uninteresting regions (grey tiles). We excluded $\roadSet$ from $\candidate$ for ease of implementation. We set $T_d=8$ and $T_c=4$.

We use Crazyflie2.1 quadrotors \cite{crazyflie} as the sensors. 
Each Crazyflie is equipped with one IR-reflective marker detected by an OptiTrack motion capture system running at $60$ Hz. 
We use Turtlebot4 robots~\cite{turtlebot}  as stations. Each Turtlebot is equipped with two IR-reflective markers and two wireless chargers.
We use a combination of Crazyswarm2 \cite{crazyswarm} and custom ROS2 packages to track and control the motion of the robots over WiFi.
All computations were performed in an Ubuntu 22.04 LTS workstation with an AMD Ryzen 9 9590X 16-core CPU and 128GB of RAM.

We use Algorithm~\ref{algo:high_level} to deploy the team and solve Problem~\ref{prob_st:main}. 
During each epoch, the high-level multi-armed bandit planner identifies six epoch goals ($\nCellsToInvestigateAtEpoch=6$). 
Then, the low-level planner coordinates the motion of the drones and ground stations to visit these cells as soon as possible. 
After the sensors visit all epoch goals and return to the stations for charging, we use the collected data to compute the next epoch goals as well as update the keep and reject sets, until the termination criterion in Algorithm~\ref{algo:high_level} was met. 

At each unclassified cell along a sensor's path, the sensor captures $30$ images using its onboard camera (Crazyflie's AI deck) and transmits them to the central computer via WiFi. 
Using OpenCV~\cite{opencv_library}, the computer counts the number of green pixels in each image to obtain noisy measurements of whether the visited cell is interesting or not.
Thus, the sensor collects $B=30$ realizations of $\nu_l$ at each visited cell $l\in\candidate\setminus(\keepSetI\cup\rejectionSetI)$.
We also performed an additional experiment with a \emph{degraded sensor}, where we introduced an additional $5\%$ classification noise to the measurements
to mimic a poor-quality sensor. 

\begin{figure}
    \centering
    \includegraphics[width=1\linewidth]{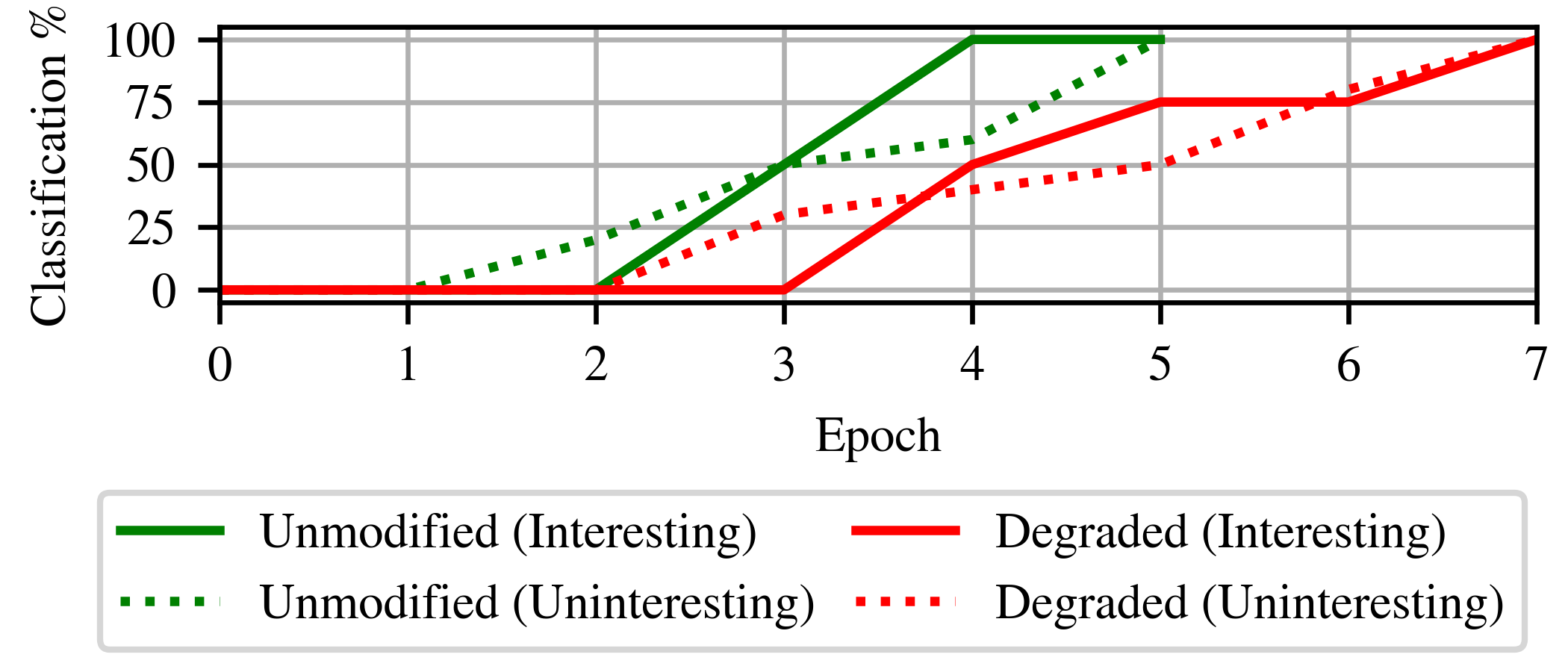}
    \caption{Classification progress of Algorithm~\ref{algo:high_level} in the hardware experiments with the unmodified and degraded sensors.}
    \label{fig:progress}
\end{figure}

Figure~\ref{fig:progress}  shows the classification progress during the hardware experiments (the percentage of the classified cells) over the course of Algorithm~\ref{algo:high_level} with the two sensors. 
For the unmodified sensor, Algorithm~\ref{algo:high_level} classifies interesting cells faster than uninteresting cells despite the sparser spatial distribution of the interesting cells. For the degraded sensor, Algorithm~\ref{algo:high_level} takes more epochs to complete the classification, as expected. However, even with the degraded sensors, Algorithm~\ref{algo:high_level} correctly identifies most of the interesting cells at the start of epoch $p=5$.
Algorithm~\ref{algo:low_level_ip} required two sensing cycles for the team to visit the epoch goals for $p=3$ with the unmodified sensor and $p=5$ with the degraded sensor.
In all other epochs, Algorithm~\ref{algo:low_level_ip} was able to find paths for the team to visit all epoch goals within a single sensing cycle.

\section{Simulation-based performance analysis}

For a more extensive assessment of the performance of the proposed approach, we also perform a simulation-based analysis. First, we study the effect of sensor accuracy (variation of $\mu_l$ for a fixed threshold $\delta$), and show that our approach is robust to noise in measured data. Second, we study the effect of varying the number of epoch goals $D$ and demonstrate a trade-off between computation time as well as utilization of the sensors.
Third, we study the effect of agility on the robot teams, and empirically show that more agile sensors typically lead to faster identification of interesting cells.
Finally, we demonstrate that our approach scales reasonably for varying team sizes.
To easily generate random scenarios, we relaxed the station admissible set constraints in this section.

To study the impact of various parameters on Algorithm~\ref{algo:high_level}, we report the results from $M\in\bbn$ Monte-Carlo simulations.
Unless specified otherwise, we consider a search environment $10\times10$ grid $\calg$ with randomly chosen obstacle set $\obstacleSet$ with $|\obstacleSet|=16$, no station admissible set restriction with $\roadSet=\calg\setminus\obstacleSet$, and  $10$ randomly chosen cells in $\candidate=\calg\setminus\obstacleSet$ chosen as interesting.
We eliminated randomly generated scenarios where an interesting cell is rendered inaccessible by the realization of the obstacle locations.
Additionally, we randomly choose $\mu_l$, the sensor accuracy at each cell $l\in\candidate$, from a uniform distribution defined over the interval $[\mu_{l,\text{interest}}^\text{worst},1]$ for every interesting cell.
Similarly, we drew $\mu_l$ from a uniform distribution defined over the interval $[0, 1 - \mu_{l,\text{interest}}^\text{worst}]$ for every uninteresting cell. 
Unless specified otherwise, we choose $\mu_{l,\text{interest}}^\text{worst}=0.8$, and fixed the threshold $\theta=0.5$ to define $\cals_{\theta}$.
For the search team, we consider $N_d=10$ sensors and $N_c=5$ stations, choose the number of epoch goals generated $\nCellsToInvestigateAtEpoch=8$ and the sample batch size $B=10$, and impose move constraints on the sensors and stations within each epoch with $T_d=8$ and $T_c=4$.

\begin{table}[t]
    \centering
    \caption{Median computation time needed to solve \eqref{prob:ip} at each epoch and the epochs needed for classification when varying the worst-case sensor accuracy $\mu_{l,\text{interest}}^\text{worst}$ (along with $0.1$, $0.9$ quantiles). The epochs needed decrease with increasing sensor accuracy.}
    \begin{adjustbox}{max width=1\linewidth}
    \begin{tabular}{cccc}
        \toprule
        Worst-case sensor & Computation
          & \multicolumn{2}{c}{Epochs needed to classify} \\\cline{3-4}
accuracy $\mu_{l,\text{interest}}^\text{worst}$ &  time (s) per epoch &  All interesting cells & All cells \\
        \midrule
        0.6 & \ 9.70 (4.83, 487.21) & \ 74 (15, 141)  & 116 (79, 141) \\
        0.8 & 11.37 (6.46, 387.38) & 17 (12, 21)  & 26 (23, 30) \\
        1.0 & 11.72 (6.18, 488.06) & 9 (7, 12)  & 15 (12, 19) \\
        \bottomrule
        \end{tabular}
    \end{adjustbox}
\label{tab:sensor_values}
\end{table}

\begin{figure}[t]
    \centering
    \includegraphics[width=0.95\linewidth,trim={0 5 0 0}, clip]{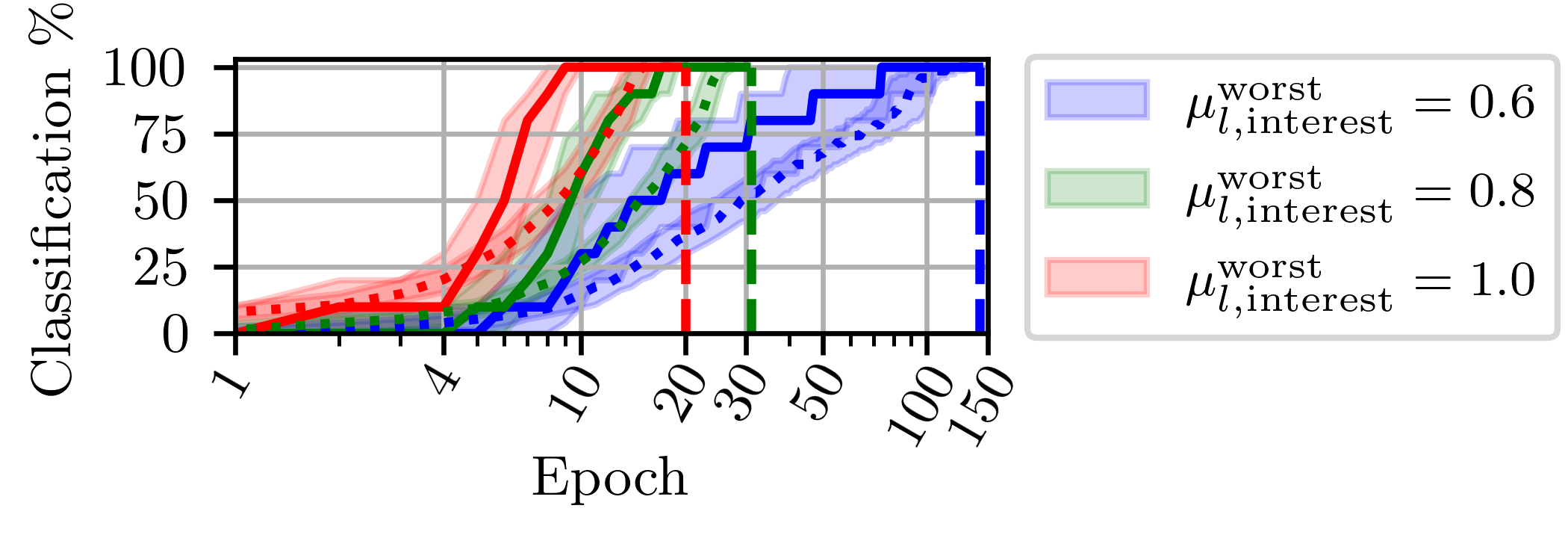}
    \caption{Classification progress of Algorithm~\ref{algo:high_level} when varying worst-case sensor accuracy $\mu_{l,\text{interest}}^\text{worst}$. Solid and dashed lines represent the classification progress for interesting and uninteresting cells, respectively. Epoch (x-axis) is in log-scale.
}
    \label{fig:sensor_values}
\end{figure}

\subsubsection{Sensor accuracy (varying the support of $\mu_l$)}

We study the effects of sensor accuracy by varying $\mu_{l,\text{interest}}^\text{worst}\in\{0.6,0.8,1.0\}$ with $M=100$. 
Specifically, we study the performance of the method under varying noise levels.
Algorithm~\ref{algo:high_level} is not aware of $\mu_{l,\text{interest}}^\text{worst}$ or $\mu_l$ for any $l\in\candidate$, but instead uses data to build the necessary  confidence intervals to address Problem~\ref{prob_st:main}.
Consequently, it has no initial benefit even when a perfect sensor is used ($\mu_{l,\text{interest}}^\text{worst}=1.0$).

Figure \ref{fig:sensor_values} shows that a team with more accurate sensor (higher $\mu_l$) completes the spatial classification problem faster.
We observe similar trends in the epochs needed for classification in Table \ref{tab:sensor_values}.
These observations are consistent with Proposition~\ref{prop:finite_time}, since a higher $\mu_{l,\text{interest}}^\text{worst}$ corresponds to a higher $\Delta_l$ by construction, which in turn decreases $P^\text{max}$, the upper bound on the termination time of Algorithm~\ref{algo:high_level}.
Finally, we observe empirically that Algorithm~\ref{algo:high_level} classifies most of the interesting cells before classifying all the uninteresting cells, which we attribute to the use of a bandit-based high-level planner in Algorithm~\ref{algo:high_level}.

\subsubsection{Number of epoch goals $D$}
Table~\ref{tab:bandit_ablation} reports the computation time and the epochs needed for classification when varying $\nCellsToInvestigateAtEpoch$ with $M=50$. 
Recall that increasing $\nCellsToInvestigateAtEpoch$ 
reduces the benefit of optimizing cell visits in the high-level planner.
Table~\ref{tab:bandit_ablation} shows that increasing $\nCellsToInvestigateAtEpoch$ typically increases the burden on the low-level planner while decreasing the epochs needed for classification.

\begin{table}[t]
    \centering
    \caption{Median computation time needed to solve \eqref{prob:ip} at each epoch and the epochs needed for classification when varying the number of epoch goals $D$ (along with $0.1$, $0.9$ quantiles). Increasing $D$ results in a trade-off --- the computation time increases, while the epochs needed decrease.}
    \label{tab:bandit_ablation}
    \begin{adjustbox}{max width=1\linewidth}
        \begin{tabular}{cccc}
            \toprule
            Number of & Computation 
             &\multicolumn{2}{c}{Epochs needed to classify} \\\cline{3-4}
epoch goals $D$  & time (s) per epoch &  All interesting cells & All cells \\
            \midrule
            5 & \ \ 9.18 (5.87, 3354.36) & 19 (12, 25)  & 29 (25, 36) \\
            8 & 11.37 (6.46, 387.38) & 17 (12, 21)  & 26 (23, 30) \\
            10 & 12.59 (5.83, 324.19) & 15 (12, 21)  & 25 (23, 28) \\
            15 & 15.71 (6.57, 594.75) & 16 (12, 25)  & 23 (20, 28) \\
            \bottomrule
            \end{tabular}
    \end{adjustbox}
\end{table}

\subsubsection{Agility of the team}

Table~\ref{tab:td_by_tc} reports the computation time and the epochs needed for classification when varying $T_d$ and $T_c$ with $M=50$. $T_d$ and $T_c$ constraint the motion of the team and arise from the energy limitations of the sensors as well as the gap between the agility of the sensors and stations.
As expected, sensors and stations that are more agile (higher $T_d$ and $T_c$) typically require lower computation time, possibly because the low-level planner has a simpler integer program to solve.
Additionally, the epochs needed also decrease with more agile sensors since the team can now cover larger parts of the search environment at each epoch.
\begin{table}[!h]
    \centering
    \caption{Median computation time needed to solve \eqref{prob:ip} at each epoch and the epochs needed for classification when varying the motion constraints on the team $T_d$ and $T_c$ (along with $0.1$, $0.9$ quantiles). Sensors and stations with more agility (higher $T_d$, $T_c$) typically need fewer epochs.}
    \label{tab:td_by_tc}
    \begin{adjustbox}{max width=1\linewidth}
        \begin{tabular}{ccccc}
            \toprule
            Sensing cycle & \multirow{2}*{$\gamma_{dc}$} & Computation 
             &\multicolumn{2}{c}{Epochs needed to classify} \\\cline{4-5}
($T_d,T_c$) & &  time (s) per epoch &  All interesting cells & All cells \\
\midrule
(8, 4) & 2 & 11.38 (6.54, 389.24) & 17 (12, 21)  & 26 (23, 30) \\
(12, 4) & 3 & 14.26 (8.97, 835.76) & 14 (10, 23)  & 22 (17, 26) \\
(16, 4) & 4 & \ 19.41 (11.85, 243.15) & 12.5 (8, 19)  & 19 (15, 23) \\
\midrule
(12, 6) & 2 & 13.57 (8.82, 194.83) & \ 15 (10, 20)  & 21 (19, 25) \\
(18, 6) & 3 & 20.45 (12.46, 65.20) & 12 (8, 17)  & 17 (14, 21) \\
(24, 6) & 4 & 27.95 (15.73, 79.49) & 10 (7, 15)  & 15 (12, 18) \\
\midrule
(16, 8) & 2 & 18.54 (11.64, 50.58) & 14 (8, 19)  & 18 (16, 23) \\
(24, 8) & 3 & 27.43 (16.23, 86.60) & 10 (5, 14)  & 14 (12, 20) \\
(32, 8) & 4 & 35.56 (21.05, 85.38) & \ 9 (5, 13)  & 13 (10, 17) \\
\bottomrule
\end{tabular}
    \end{adjustbox}
\end{table}

\subsubsection{Scalability}

Table~\ref{tab:scalability} reports the computation time and the epochs needed for classification when varying team sizes  with $M=50$. 
Increasing the number of sensors $N_d$ from $5$ to $14$ almost halves the epochs needed, but nearly doubles the computation time per epoch.
Table~\ref{tab:scalability} also provides evidence for the multi-agent coordination.
However, even for moderately sized teams, the $0.9$-quantile computation time per epoch for Algorithm~\ref{algo:high_level} is less than $10$ minutes.

\begin{remark}
    We refer the reader to our prior work~\cite{Siddharth2024data} for simulation-based validation on larger and more complex environments, including a $20\times 20$ grid $\calg$. For larger problems where \eqref{prob:ip} becomes computationally expensive,~\cite{Siddharth2024data} also discusses a more scalable heuristic for the low-level planner.
\end{remark}

\begin{table}[t]
    \centering
    \caption{Median computation time needed to solve \eqref{prob:ip} at each epoch and the epochs needed for classification when varying the team sizes (along with $0.1$, $0.9$ quantiles). Larger teams need fewer epochs but need more compute time.}\label{tab:scalability}
    \begin{adjustbox}{max width=1\linewidth}
        \begin{tabular}{cccc}
            \toprule
            Team size & Computation
             &\multicolumn{2}{c}{Epochs needed to classify} \\\cline{3-4}
($N_d,N_c$)  &  time (s) per epoch &  All interesting cells & All cells \\
\midrule
\ (5, 5) & \ \ 6.17 (3.77, 192.55) & 23 (17, 33)  & 41 (36, 46) \\
\ (7, 7) & \ \ 8.07 (5.73, 140.56) & 22 (13, 30)  & 36 (32, 42) \\
(10, 5) & 11.37 (6.46, 387.38)     & 17 (12, 21)  & 26 (23, 30) \\
\ (10, 10) & 12.74 (8.74, 525.27)  & 18 (13, 27)  & 30 (25, 36) \\
(14, 7) & \ 15.73 (10.38, 705.02)  & 13 (10, 17)  & 21 (20, 29) \\
\bottomrule
\end{tabular}
\end{adjustbox}
\end{table}

\subsubsection{Comparison with baselines}
\begin{figure}[t]
    \centering
    \includegraphics[width=0.95\linewidth,trim={0 5 0 0}, clip]{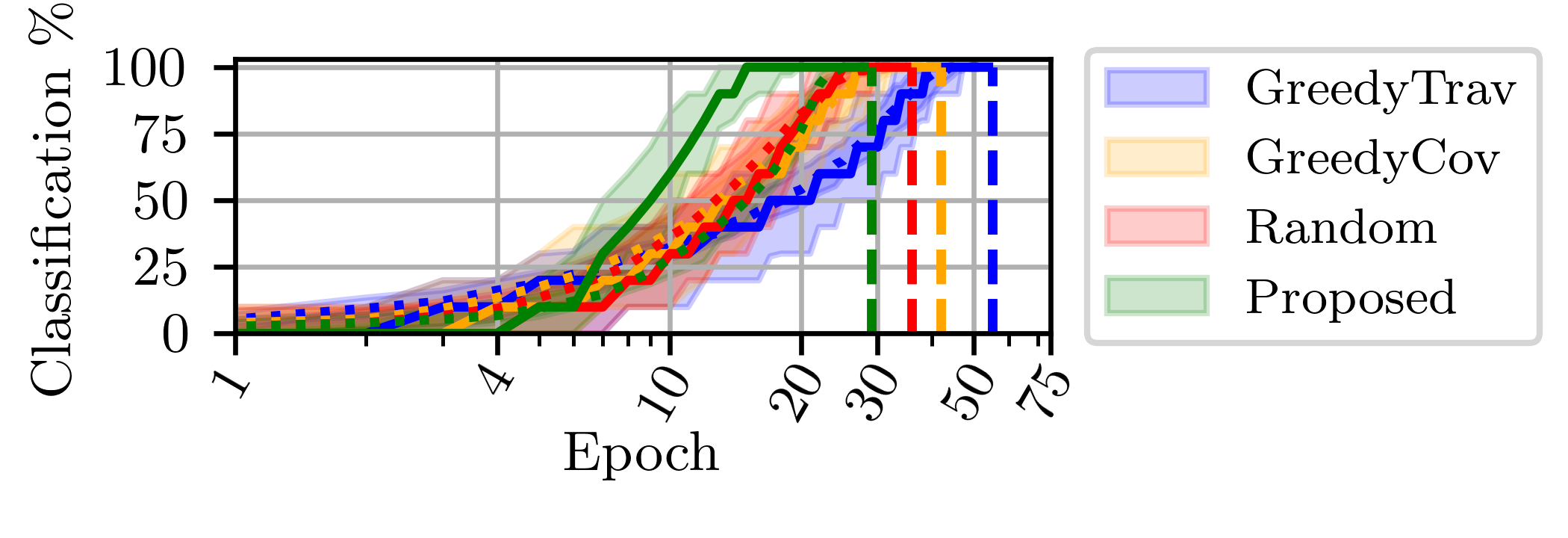}
    \caption{Comparison of the classification progress of Algorithm~\ref{algo:high_level} and three baselines --- \texttt{Random}, \texttt{GreedyCov}, and \texttt{GreedyTrav}. Solid and dashed lines represent the classification progress for interesting and uninteresting cells.}
    \label{fig:baselines}
\end{figure}

Figure~\ref{fig:baselines} shows that Algorithm~\ref{algo:high_level} completes the spatial classification task and identifies all interesting cells faster than three \emph{data-agnostic} baselines for $D=N_d=10$. The baselines differ from Algorithm~\ref{algo:high_level} on how they choose $\cale_p$ in Step~\ref{step:high_level_epoch_goal} --- \texttt{Random} chooses $\cale_p$ randomly from the unclassified cells, and \texttt{GreedyCov}/\texttt{GreedyTrav} greedily maximize coverage/minimize team travel distance. 

\vspace*{-1.0em}
\section{Conclusion}

We presented an iterative algorithm to address the spatial classification problem using constrained robots. We used a combination of multi-arm bandits and optimization-based planning to design the proposed data-driven algorithm.
Using hardware and simulation experiments, we demonstrated the efficacy of our approach in a variety of settings. 
Our future work will focus on improving the high-level planner using constraints from the low-level planner, and deriving upper bounds on the regret for the high-level planner.
Additionally, we will investigate spatio-temporal monitoring of dynamic environments, which may require incorporation of periodic re-initialization and adaptation into the proposed approach.

\vspace*{-0.5em}
\section{Acknowledgements}
We thank Dr. Parth Thaker for invaluable discussions during the development of this work.

\bibliography{references.bib,IFACbib.bib}

\appendix

\subsection{Proof of Proposition~\ref{prop:anytime}}
\label{app:anytime}

Consider two events ---   $\cale_\keepSet{p}=\left\{{{\keepSet{p}\setminus\cals_{\theta-\epsilon}}\neq \emptyset}\right\}$ and $\cale_\rejectionSet{p}=\left\{{\rejectionSet{p}\cap\cals_{\theta+\epsilon} \neq \emptyset}\right\}$.
Specifically, at any epoch $p$, $\cale_\keepSet{p}$ and $\cale_\rejectionSet{p}$ are the undesirable events that a cell in $\candidate\setminus\cals_{\theta-\epsilon}$ is included in the keep set $\keepSet{p}$ and a cell in $\cals_{\theta+\epsilon}$ is included in the reject set $\rejectionSet{p}$ respectively.
We need to show that $\mathbb{P}\left[{
\left({\bigcup_{p\geq 1}}\cale_\keepSet{p}\right) 
\bigcup
\left({\bigcup_{p\geq 1}}\cale_\rejectionSet{p}\right)
}\right] \leq \delta$.

By Boole's inequality, it is suffices to show that $\mathbb{P}\left[{
{\bigcup_{p\geq 1}}\cale_\keepSet{p}}\right]\leq \delta/2$ and $\mathbb{P}\left[{
{\bigcup_{p\geq 1}}\cale_\rejectionSet{p}}\right]\leq \delta/2$. 
Observe that $U_{l}(p, \delta)$ in \eqref{eq:acq_fun_U} is the upper confidence bound in~\cite[Eq. (2)]{pmlr-v51-jun16} with $\delta$ replaced with $\delta/(2|\candidate|)$.
By \eqref{eq:acq_fun_U} and~\cite[Lem. 1]{pmlr-v51-jun16}, 
\begingroup\makeatletter\def\f@size{9}\check@mathfonts
\begin{align}
    \mathbb{P}\left[\bigcap\nolimits_{p\geq 1}\left\{\left|\mu_l - \hat{\mu}_l\right| \leq  U_{l}(p, \delta)\right\} \right] \geq 1-\frac{\delta}{2|\candidate|},\label{eq:pmlr_result}
\end{align}
\endgroup
for any cell $l\in\candidate$. 
Recall that the event  $\cale_\keepSet{p}$ occurs at some $p$ if and only there is some cell $l\in\candidate$ such that $\mu_l < \theta - \epsilon \leq \hat{\mu}_{l}(p)- U_{l}\left(p,\delta\right)$ at $p$. 
Using \eqref{eq:pmlr_result} and Boole's inequality,
\begingroup\makeatletter\def\f@size{9}\check@mathfonts
\begin{align}
    \mathbb{P}\left[{
{\bigcup\nolimits_{p\geq 1}}\cale_\keepSet{p}}\right] 
&\leq\mathbb{P}\left[\bigcup\nolimits_{l\in\candidate}\bigcup\nolimits_{p\geq 1}\left\{{|\hat{\mu}_{l}(p)-\mu_l|\geq U_{l}\left(p,\delta\right)}\right\} \right] \nonumber\\
&\leq \sum\nolimits_{l\in \candidate}
    \frac{\delta}{2|\candidate|} \leq \frac{\delta}{2}.\label{eq:pmlr_result_all_cells}
\end{align}%
\endgroup
The proof for $\mathbb{P}\left[{
{\bigcup_{p\geq 1}}\cale_\rejectionSet{p}}\right]\leq \delta/2$ follows similarly.\hfill$\blacksquare$

\subsection{Proof of Proposition~\ref{prop:finite_time}}
\label{app:finite_time}

We simplify the analysis by considering the data collected only at $\cale_p$ at each epoch $p$ and ignore the effect of data collected along the way on the classification.
Due to the monotonicity of $U_l$ and $|\calh_l|$, such an analysis holds even when Algorithm~\ref{algo:high_level} uses data collected along the way.
Consequently, we study the corresponding bandit problem $(\candidate,\{\mu_l\}_{l\in\candidate},\theta,\delta,\epsilon)$~\cite{pmlr-v51-jun16}, and ignore the effect of the sensor paths on the classification. 

\emph{Claim 1:} For every $l\in\candidate$, consider any epoch $p_l\in\bbn$ such that $U_l(p_l,\delta)\leq\Delta_l/2$. Then,  with probability $1-\delta$, $l$ is assigned either to $\keepSetI$ or $\rejectionSetI$ by epoch $p_l$.\newline
\emph{Proof of Claim 1:} From \eqref{eq:update_sets}, $l$ is assigned to either $\keepSetI$ or $\rejectionSetI$ by epoch $p$ if $|\hat{\mu}_l(p) - \theta|\geq U_l(p,\delta) - \epsilon$.
Using triangle inequality, $U_l(p_l,\delta)\leq\Delta_l/2$, \eqref{eq:delta_defn}, and \eqref{eq:pmlr_result_all_cells}, $|\hat{\mu}_l(p_l) - \theta|\geq U_l(p_l,\delta) -\epsilon$, and the proof is completed using \eqref{eq:update_sets}.\qed

We define $P_l\in\bbn$ as the number of times Algorithm~\ref{algo:high_level} directs the team to visit cell $l$.
For every cell $l$ and any epoch $p_l$ that satisfies Claim 1, $|\calh_l(p_l)|=P_lB$ for sample batch size $B$ by Step~\ref{step:high_level_epoch_goal} of Algorithm~\ref{algo:high_level}.
We obtain $P_l$ in \eqref{eq:P_l_defn} using $U_l(p_l,\delta)\leq\Delta_l/2$ and~\cite[Eq. (6)]{pmlr-v51-jun16}. 

\emph{Claim 2}: With probability $1-\delta$, Algorithm~\ref{algo:high_level} terminates by $
    P^\text{max}=\frac{1}{\nCellsToInvestigateAtEpoch}\sum_{l\in\candidate\setminus
\scrd_\Delta} P_l + \max_{l\in\scrd_\Delta} P_l
$ epochs.
\newline
\emph{Proof of Claim 2:}
We split the progress of Algorithm~\ref{algo:high_level} into two phases --- Phase 1, the initial phase, where all epochs has more than $D$ unclassified cells, and otherwise as Phase 2. 
Let $\scrd\subset\candidate$, $|\scrd|=D$ denote the set of unclassified cells at the start of Phase 2.
Then, using the definition of $P_l$, Algorithm~\ref{algo:high_level} terminates within $P^\text{max}$ epochs with probability $1-\delta$, 
\begingroup\makeatletter\def\f@size{9}\check@mathfonts
\begin{align}
    P^\text{max}(\scrd)=\frac{1}{\nCellsToInvestigateAtEpoch}\sum\nolimits_{l\in\candidate\setminus
\scrd} P_l + \max\nolimits_{l\in\scrd} P_l.\label{eq:P_l_construction}
\end{align}
\endgroup
Here, \eqref{eq:P_l_construction} uses the observations that the low-level planner coordinates the team to visit each epoch goal at least once, Phase 1 involves at most $\sum_{l\in\candidate\setminus\scrd} P_l$ visits with $|\cale_p|=D$, and Phase 2 has $\cale_p=\candidate\setminus(\keepSetI\cup\rejectionSetI)$ with $|\cale_p|\leq D$.
We observe that $\scrd_\Delta$ maximizes \eqref{eq:P_l_construction}, which completes the proof.\qed

By Proposition~\ref{prop:anytime}, Algorithm~\ref{algo:high_level} satisfies the labeling criterion \eqref{eq:labeling_error} whenever it terminates, which completes the proof.
$\hfill\blacksquare$
\end{document}